# Expert Mind: A Retrieval-Augmented Architecture for Expert Knowledge Preservation in the Energy Sector

**Diego Ezequiel Cervera**

*Independent Researcher*

Buenos Aires, Argentina

cervera.diego@gmail.com

## Abstract

The departure of subject-matter experts from industrial organizations results in the irreversible loss of tacit knowledge that is rarely captured through conventional documentation practices. This paper proposes Expert Mind, an experimental system that leverages Retrieval-Augmented Generation (RAG), large language models (LLMs), and multimodal capture techniques to preserve, structure, and make queryable the deep expertise of organizational knowledge holders. Drawing on the specific context of the energy sector, where decades of operational experience risk being lost to an aging workforce, we describe the system architecture, processing pipeline, ethical framework, and evaluation methodology. The proposed system addresses the knowledge elicitation problem through structured interviews, think-aloud sessions, and text corpus ingestion, which are subsequently embedded into a vector store and queried through a conversational interface. Preliminary design considerations suggest Expert Mind can significantly reduce knowledge transfer latency and improve onboarding efficiency. Ethical dimensions including informed consent, intellectual property, and the right to erasure are addressed as first-class design constraints.

## I. INTRODUCTION

The departure of experienced professionals from complex industrial organizations represents one of the most underestimated risks in knowledge management. In domains such as oil and gas, power generation, and petrochemical processing, decades of operational knowledge are often stored exclusively in the minds of senior engineers and technicians. When these individuals retire or transition to other roles, the organizations they leave behind face a compounded challenge: not merely the loss of a skilled worker, but the permanent erasure of contextual judgment, heuristic reasoning, and institutional memory that cannot be recovered through documentation alone.

Traditional knowledge management approaches, including procedure manuals, standard operating procedures, and shadowing programs, address only the explicit dimension of organizational knowledge. Polanyi's seminal distinction between explicit and tacit knowledge [1] remains as relevant today as when it was first articulated: the knowledge that experts *know how* to do is fundamentally different from the knowledge they can articulate on demand. This tacit dimension, encompassing pattern recognition, diagnostic intuition, and experiential judgment, has historically resisted systematic capture.

Recent advances in large language models (LLMs), vector-based semantic search, and multimodal processing have opened a viable technical pathway toward tacit knowledge preservation. The Retrieval-Augmented Generation (RAG) paradigm [2] combines the generative capacity of LLMs with structured retrieval from domain-specific corpora, enabling systems to produce contextually grounded responses from curated knowledge bases. When applied to the problem of expert knowledge preservation, RAG provides a compelling architecture for converting interview transcripts, technical documents, and think-aloud recordings into a queryable conversational agent.

This paper presents Expert Mind, an experimental system designed to preserve the tacit knowledge of subject-matter experts through a four-layer architecture: multimodal knowledge capture, LLM-driven processing and indexing, vector store persistence, and a conversational query interface. The system is contextualized within the energy sector, where the knowledge crisis is most acute, though the architecture is domain-agnostic.

The contributions of this paper are threefold: (1) a concrete system architecture for expert knowledge preservation using RAG and multimodal capture; (2) an ethical framework addressing the unique considerations of digitizing an individual's cognitive profile; and (3) a proposed evaluation methodology for measuring knowledge fidelity and organizational impact.

The remainder of this paper is organized as follows. Section II reviews related work in knowledge management, expert systems, and conversational AI. Section III describes the proposed system architecture. Section IV addresses ethical and legal considerations. Section V outlines the evaluation methodology and expected outcomes. Section VI concludes with directions for future research.

## II. LITERATURE REVIEW

### A. Tacit Knowledge and the Knowledge Elicitation Problem

The concept of tacit knowledge, first introduced by Polanyi [1] and later operationalized within knowledge management by Nonaka and Takeuchi [3], describes knowledge that is difficult to transfer through verbalization or written documentation. Within the SECI model [3], socialization and externalization are identified as the primary conversion modes through which tacit knowledge becomes accessible to others. Expert Mind targets the externalization quadrant by converting spoken and written expert output into structured, retrievable knowledge artifacts.

Knowledge elicitation research has long grappled with the challenge of surfacing tacit expertise. Protocol analysis and think-aloud methods [4] have proven effective in capturing domain reasoning by prompting experts to verbalize their thought processes during problem-solving. More recently, structured interviewing techniques adapted from cognitive task analysis [5] have been employed in high-stakes domains including aviation, military operations, and nuclear energy. Expert Mind integrates both approaches as primary capture modalities.

### B. Retrieval-Augmented Generation

Lewis et al. [2] introduced Retrieval-Augmented Generation as a method for enhancing generative models with non-parametric memory retrieved from external document stores. Unlike purely parametric models, RAG systems decouple knowledge storage from language generation, enabling continuous knowledge updates without retraining. This property is particularly valuable for organizational knowledge systems, where corpora evolve as new expertise is captured.

Subsequent work has expanded the RAG paradigm to multimodal inputs [6], enabling retrieval from audio transcripts, video annotations, and structured documents. The integration of dense passage retrieval [7] with transformer-based generation has demonstrated strong performance on open-domain question answering tasks, providing a foundation for domain-specific expert systems.

### C. Expert Systems and Digital Twins

Early expert systems [8] relied on hand-crafted rule bases elicited from domain experts and encoded by knowledge engineers. While effective in narrow domains, these systems were brittle, expensive to maintain, and required continuous expert involvement. Modern LLM-based approaches obviate much of the knowledge engineering burden by learning distributional representations directly from text.

The concept of a digital twin [9], originally applied to physical systems in industrial IoT, has been extended to human cognitive profiles. Saddik [10] proposed digital twins of humans as socially engaging entities that replicate behavioral and cognitive patterns. Expert Mind draws on this lineage while grounding the architecture in practical knowledge management objectives rather than social simulation.

### D. Knowledge Management in the Energy Sector

The energy sector faces a documented "crew change" challenge [11], wherein a disproportionate share of experienced petroleum engineers, geoscientists, and operations specialists are approaching retirement simultaneously. Studies estimate that up to 50% of the experienced workforce in major oil and gas companies may retire within the next decade [12], creating acute pressure on knowledge transfer programs. Existing digital knowledge management platforms have shown limited success in capturing the heuristic judgment central to reservoir engineering, process optimization, and equipment diagnostics.

## III. PROPOSED SYSTEM ARCHITECTURE

Expert Mind is organized around four functional layers that process expert knowledge from raw multimodal input to structured, queryable output. Fig. 1 provides a schematic overview of the system pipeline.

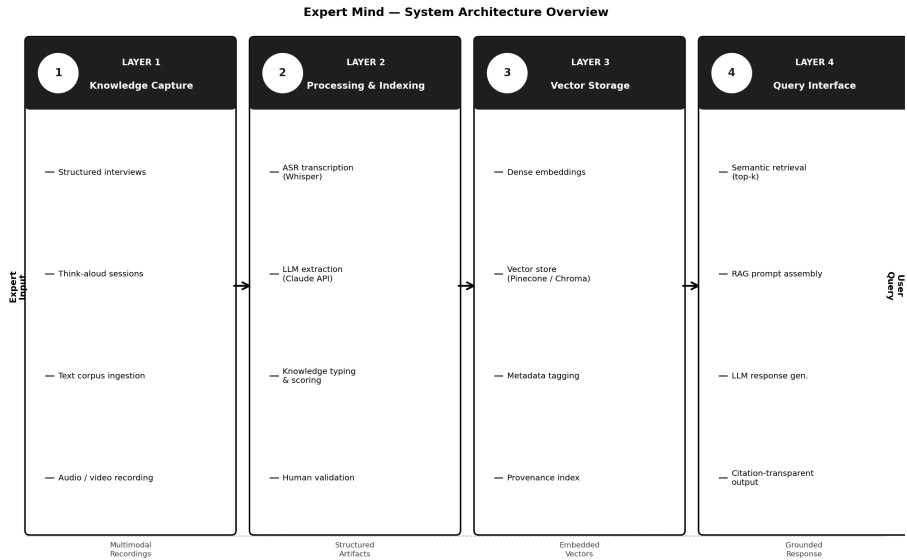

Fig. 1. Four-layer architecture of the Expert Mind knowledge preservation system.

### A. Layer 1: Multimodal Knowledge Capture

The capture layer is responsible for eliciting and recording expert knowledge through three complementary modalities. Structured interviews, conducted following a cognitive task analysis protocol, prompt experts to articulate decision rationale for representative scenarios drawn from their domain experience. Sessions are conducted in 60–90 minute blocks and recorded in high-definition audio and video.

Think-aloud sessions complement structured interviews by observing experts as they navigate real or simulated operational problems. This modality is particularly effective for surfacing procedural and diagnostic knowledge that experts may not spontaneously verbalize in interview contexts.

Finally, a corpus ingestion pipeline processes pre-existing textual artifacts: technical reports, email correspondence, annotated procedures, and internal knowledge base entries. Natural language processing filters are applied to remove personally identifiable information and normalize formatting prior to downstream processing.

### B. Layer 2: Processing and Knowledge Extraction

Audio and video recordings are transcribed using OpenAI Whisper [13], a transformer-based automatic speech recognition (ASR) model that achieves state-of-the-art performance on domain-specific speech in both English and Spanish. Transcripts are subsequently processed by a Claude API [14] extraction pipeline that identifies and structures four knowledge artifact types: (1) factual claims, (2) decision criteria and heuristics, (3) anomaly recognition patterns, and (4) best practices and lessons learned.

Each extracted artifact is assigned a confidence score based on corroboration across multiple source documents and validated by the originating expert before ingestion into the knowledge base. This human-in-the-loop validation step is essential for maintaining knowledge fidelity and preventing hallucination propagation.

### C. Layer 3: Vector Store Persistence

Validated knowledge artifacts are embedded using a dense text encoder and stored in a vector database (Pinecone [15] for production; ChromaDB for prototyping). Each vector record is augmented with structured metadata: source document identifier, capture date, artifact type, confidence score, and domain tag. This metadata layer enables filtered retrieval, supporting queries that target specific knowledge types or temporal ranges.

### D. Layer 4: Conversational Query Interface

The query interface exposes a RESTful API and a web-based chat frontend through which authorized users can pose natural language questions. Incoming queries are embedded and used to retrieve the top-k most semantically similar knowledge artifacts from the vector store. Retrieved artifacts are assembled into a structured prompt context and passed to an LLM (Claude API) that generates a grounded response, explicitly citing source artifacts.

| Layer | Component | Primary Tool | Alternative |
|---|---|---|---|
| Capture | Transcription | OpenAI Whisper | Azure Speech |
| Processing | LLM Extraction | Claude API | GPT-4 |
| Storage | Vector DB | Pinecone | ChromaDB |
| Storage | File Store | Azure Blob | AWS S3 |
| Query | Orchestration | LlamaIndex | LangChain |
| Query | Frontend | React + Tailwind | Vue.js |
| Query | Backend | FastAPI (Python) | Node.js |

The system implements a citation-transparent response mode in which every factual claim in the generated response is attributed to a specific knowledge artifact, enabling users to trace the provenance of each answer and assess its reliability. This design choice directly addresses the risk of confabulation in LLM-generated responses.

*E. Technology Stack*

Table I summarizes the selected technologies for each system layer, including recommended alternatives for deployments with differing infrastructure constraints.

**TABLE I**
*Technology Stack by System Layer*

## IV. ETHICAL AND LEGAL CONSIDERATIONS

The digitization of an individual's cognitive profile raises ethical questions that are qualitatively distinct from those associated with conventional data systems. Expert Mind treats ethical compliance not as an afterthought but as a first-class architectural constraint, instantiated in system design decisions rather than external policy overlays.

*A. Informed Consent and Autonomy*

All knowledge capture activities require prior written informed consent from the participating expert. Consent documentation specifies: (1) the scope of knowledge to be captured, (2) the intended uses of the resulting digital twin, (3) the organizations and individuals who will have query access, and (4) the duration of knowledge retention. Participation is strictly voluntary, and withdrawal rights are preserved throughout the engagement.

*B. Intellectual Property*

The ownership of expert knowledge artifacts presents a novel legal question at the intersection of employment law and intellectual property. Expert Mind recommends that participating organizations and experts execute a knowledge licensing agreement prior to capture activities, explicitly addressing whether knowledge artifacts are considered works-for-hire, licensed contributions, or jointly owned intellectual property.

*C. Bias and Epistemic Limitations*

Expert knowledge is inherently situated: it reflects the experience, assumptions, and cognitive biases of a specific individual operating within a specific organizational and historical context. Expert Mind implements a mandatory disclosure mechanism that surfaces the provenance and confidence metadata of every retrieved artifact, ensuring that users are aware of the limitations and potential biases embedded in generated responses.

*D. Right to Erasure*

Consistent with data protection frameworks including the General Data Protection Regulation (GDPR) [16], Expert Mind provides a complete erasure pathway through which an expert may request the permanent deletion of all knowledge artifacts derived from their contributions. The system architecture ensures that erasure is propagated across the vector store, metadata index, and all derived model artifacts within a defined service-level window.

*E. Voice and Likeness*

The optional integration of voice cloning technology introduces additional ethical complexity. Voice synthesis from a named individual's recordings constitutes the creation of a synthetic identity artifact with potential for misuse. Expert Mind requires explicit written consent for voice cloning features, separate from the base knowledge capture consent, and prohibits the use of synthetic voice outputs in any context that could reasonably be mistaken for authentic speech by the originating individual.

## V. EVALUATION METHODOLOGY AND EXPECTED OUTCOMES

As a proposed experimental system, Expert Mind does not yet report empirical results. This section describes the evaluation framework that will be applied during the pilot deployment phase, along with anticipated performance benchmarks derived from comparable RAG-based knowledge systems in the literature.

## A. Knowledge Fidelity Evaluation

Knowledge fidelity measures the degree to which system-generated responses accurately reflect the expert's original knowledge. The primary evaluation instrument is an expert review protocol in which the originating expert rates a stratified random sample of system responses on a five-point accuracy scale. A secondary instrument employs domain-expert annotators blind to the system's source artifacts.

Based on performance benchmarks from comparable RAG systems applied to domain-specific question answering [17], we anticipate a baseline accuracy of 78–85% on factual claims, with the human validation layer expected to raise this to above 85% upon system maturity.

## B. Adoption and Usability Metrics

User adoption will be tracked through interaction logs collected during the 5–10 person organizational pilot. Key metrics include: weekly active query volume, query resolution rate (defined as a query for which no human follow-up is required), mean time to answer, and Net Promoter Score (NPS) collected through periodic user surveys.

## C. Organizational Impact Assessment

Organizational impact will be measured through a controlled comparison of onboarding duration and performance ramp-up between cohorts with and without access to Expert Mind. Additional indicators include: the number of operational decisions for which the system provides documented expert justification, and the reduction in senior engineer consultation time attributable to system-mediated knowledge retrieval.

## D. Expected Outcomes Summary

Table II summarizes the anticipated outcomes across evaluation dimensions, with target benchmarks and measurement instruments.

TABLE II
*Anticipated Evaluation Outcomes*

| Dimension | Metric | Target | Instrument |
|---|---|---|---|
| Knowledge Fidelity | Response accuracy | >85% | Expert review protocol |
| Knowledge Fidelity | Correction rate | <10% | Validation logs |
| Adoption | Weekly query volume | >50 queries/wk | Interaction logs |
| Adoption | Net Promoter Score | >40 | User surveys |
| Org. Impact | Onboarding reduction | >20% | HR analytics |
| Org. Impact | Consultation time saved | >15% | Time tracking |

## VI. CONCLUSION AND FUTURE WORK

This paper has presented Expert Mind, a Retrieval-Augmented Generation architecture for the preservation of tacit expert knowledge in the energy sector. By combining multimodal knowledge capture, LLM-driven extraction, vector store persistence, and a conversational query interface, the system addresses a critical organizational challenge that existing knowledge management approaches have failed to resolve.

The proposed architecture offers three primary contributions to the knowledge management and applied AI literature: a concrete, implementable system design for expert knowledge digitization; an ethical framework that treats consent, intellectual property, and the right to erasure as architectural requirements; and a rigorous evaluation methodology that enables empirical assessment of knowledge fidelity and organizational impact.

The energy sector context underscores the urgency of this work. As an aging expert workforce approaches retirement, the window for knowledge capture is narrowing. Expert Mind offers a technically feasible and ethically grounded pathway for preserving the irreplaceable operational knowledge that these professionals have accumulated.

### A. Limitations

Several limitations of the current design merit acknowledgment. First, the quality of extracted knowledge is inherently bounded by the quality of capture sessions; experts who are poor verbalizers may contribute less rich corpora. Second, the system addresses individual expert knowledge preservation but does not model the collective, distributed knowledge that emerges from team interactions. Third, the proposed evaluation methodology requires longitudinal data collection that extends the experimental timeline.

## B. Future Work

Several directions are identified for future investigation. Multi-expert fusion, in which knowledge artifacts from multiple experts are integrated into a coherent, cross-validated knowledge graph, represents a natural extension of the current architecture. Additionally, the integration of continual learning mechanisms would enable the knowledge base to evolve as captured experts acquire new experience post-digitization. Finally, the application of Expert Mind to domains beyond energy, including healthcare, legal practice, and advanced manufacturing, warrants exploration.